# Wavelets and continuous wavelet transform for autostereoscopic multiview images

VLADIMIR SAVELJEV[1,2,*]

[1]*Dept. of Inform. Display Engineering, Hanyang University, Seoul 133-791, Korea*
[2]*Imaging Media Research Center, KIST, Seoul 136-791, Korea*
*\*Corresponding author: saveljev.vv@gmail.com*

**Recently, the reference functions for the synthesis and analysis of the autostereoscopic multiview and integral images in three-dimensional displays we introduced. In the current paper, we propose the wavelets to analyze such images. The wavelets are built on the reference functions as on the scaling functions of the wavelet analysis. The continuous wavelet transform was successfully applied to the testing wireframe binary objects. The restored locations correspond to the structure of the testing wireframe binary objects. © 2015 Optical Society of America**

*OCIS codes: (100.6890) Three-dimensional image processing, (100.7410) Wavelets, (120.2040) Displays.*

http://dx.doi.org/10.1364/OL.99.099999

The synthesis and analysis of images are two main concepts of the image processing. The real-time processing of the multiview images is probably most important problem of three-dimensional (3D) imaging and image processing in autostereoscopic multiview and integral displays [1], [2].

The processing can be based on a set of functions. Such functions may be helpful for the real-time image synthesis and analysis. For instance, the computer-generated holograms can be synthesized from the hologram patterns [3], [4].

The extraction of depth is one of particular problems of the analysis of multiview and integral images [5]. The depth can be extracted using various methods including the computational integral imaging reconstruction [6], [7].

The autostereoscopic multiview images can be also synthesized from the voxel patterns [8]. The visual 3D images were confirmed, and the patterns were further developed as the reference functions [9]. The synthesis based on patterns is applicable for the real-time autostereoscopic three-dimensional displays.

Using the mentioned reference functions, the structure of multiview images can be analyzed, and locations of voxels can be extracted. Basing on that, it is possible to build a depth map, which can be effectively used in the further processing of 3D multiview images, e.g., in 3D broadcasting.

The above functions were applied for depth extraction under various conditions including noised and grayscale images and showed a good validity [10], [11]. However, the functions [9] are not orthogonal, they do not comprise a basis.

A promising technique of the real-time processing is the wavelet analysis. Generally, the wavelets can be built by various methods. One of them is to start from the scaling functions [12]. The scaling functions should satisfy certain conditions, i.e., the two-scale relation, stability (Rietz function), and the partition of the unity. Then, a wavelet can be built as a linear combination of scaling functions [13]

$$\psi(t) = \sum_{k} \beta_k \sqrt{2}\varphi(2t - k) . \qquad (1)$$

where $\psi$ is a wavelet, $\varphi$ is a scaling function, and $\beta$ is a coefficient.

A known scaling function which satisfies all necessary conditions is the Haar function [14]. The functions [9] are based on rectangular unit pulse which actually is the shifted Haar function.

The Haar function itself defines a family of wavelets [15]. Similarly, the multiview reference functions can define wavelets. The preliminary considerations were presented at the conferences [16] and more. In the current paper, we develop such wavelets and demonstrate the results of the wavelet analysis of a multiview images.

For the testing purpose, we use two binary wireframe objects: the spatial diagonals of the cube and the edges of the tetrahedron (Fig. 1). The testing object cube was proposed in [9]. The dimensions of the cube are 8x8x8. The testing object tetrahedron was proposed in [8]; its dimensions are 40x40x9, so it occupies the 3D space of approx. 15 thousands voxels.

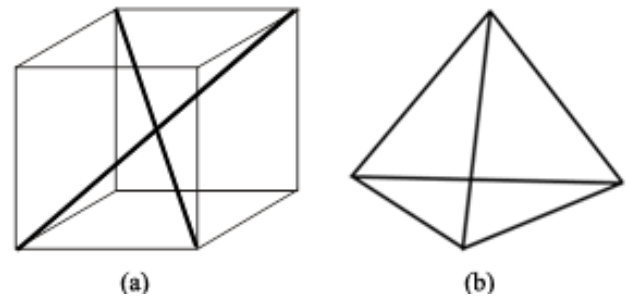

Fig. 1. Testing objects: (a) spatial diagonals of cube, (b) edges of tetrahedron.

The testing images are computer-generated multiview images of the above objects. The multiview images are built from the reference functions; the resulting computer-generated testing images are shown in Fig. 2. The testing image Fig. 2(b) additionally includes two points in the corners of the -5th plane ($z = -5$). One image cell (elemental image) of both images contains 60x60 pixels.

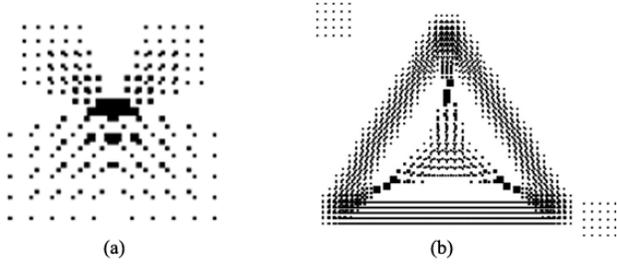

Fig. 2. Computer-generated testing multiview images of objects shown in Fig. 1.

The reference functions define the wavelets as mentioned above. According to [9], the 1D functions can be written as

$$F_k(x) = \sum_{i=0}^{|k|-1} r\left(\frac{x - x_{kj}}{w_k}\right), \quad (2)$$

where $k$ is the number of discrete plane, $x$ is the coordinate of the current point, $i$ is the running indices, $r(x)$ is the rectangular unit impulse function (the unit impulse of the unit width centered at the origin), $x_{ij}$ is the coordinate of the center of the pulse, $w_k$ is the width of the pulse. The formula (2) works in the projective space as described in [9].

The functions (2) can be used in the autostereoscopic displays with the horizontal parallax only (HPO). The graphs of functions are given in Fig. 3.

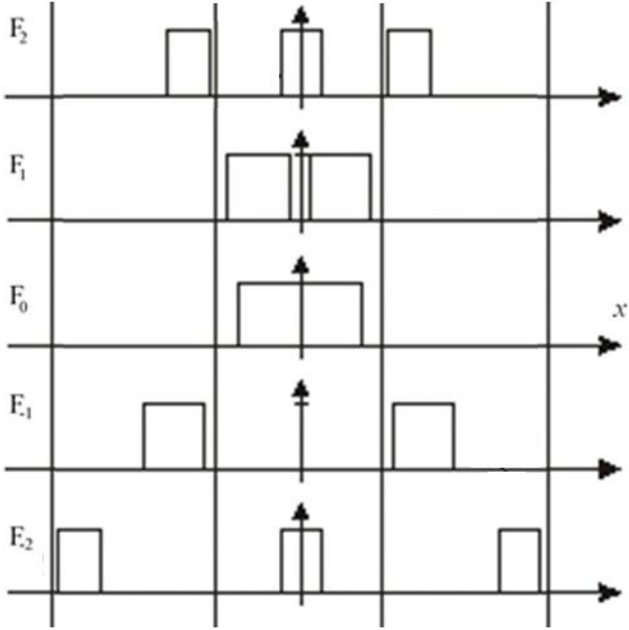

Fig. 3. 1D reference functions.

Note that for sake of convenience, the numeration of the depth planes is altered as compared to [9]. Namely, in the current paper, the index of the function is equal to the index of the reference function minus one. The sign of the index is matter of convenience. For instance, the positive functions in a barrier display become negative ones in a lenticular display.

In the case of a full-parallax display, the following two-dimensional (2D) functions can be considered,

$$F_k(x, y) = \left[\sum_{i=0}^{|k|-1} r\left(\frac{x - x_{ki}}{w_k}\right)\right] \cdot \left[\sum_{j=0}^{|k|-1} r\left(\frac{y - y_{kj}}{w_k}\right)\right], \quad (3)$$

where $y$ is the coordinate, $j$ is the index, $y_{ij}$ is the coordinate of the center of the pulse along the $y$-axis.

A 2D function is the product of two 1D functions (2) along $x$- and $y$-axes. Examples of 2D functions (3) are graphically shown in Fig. 4.

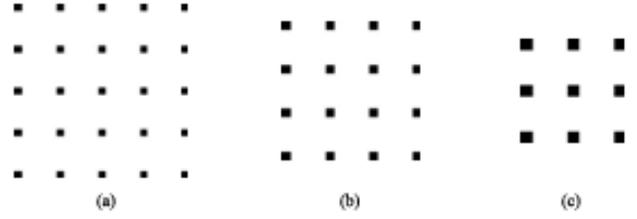

Fig. 4. 2D reference functions $-4 \leq k \leq -2$ (from left to right).

To analyze the multiview image and recognize the location of voxels in space using the functions (2), (3), we applied the convolution technique [9]. A location is considered as recognized, when the convolution at this point is higher than a predefined threshold. This way, the spatial 3D coordinates of testing objects were restored [10].

The convolution threshold is the same in all planes because of the partition of unity property of the reference functions (3). The threshold is numerically equal to the product of the cell area multiplied by the squared number of the gray levels in the image. For instance, for 256 gray levels and cells of 60 x 60 pixels, this estimation gives the threshold $2.4 \cdot 10^8$, i.e. 2.4 units (one unit equals $1 \cdot 10^8$). This is in a good coincidence with the empirically estimated threshold 1.7 units.

For testing image Fig. 2(a), the extracted locations are shown in Fig. 5 by planes.

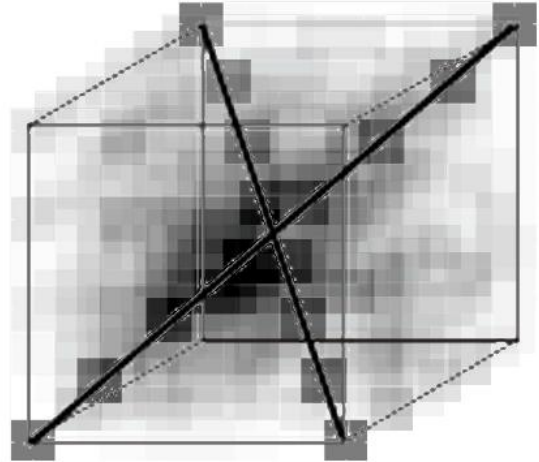

Fig. 5. The results of the convolution analysis of the spatial diagonals by discrete $x, y$ locations.

The restored spatial structure of the testing object Fig. 2(b) is shown in Fig. 6 as a color coded map of recognized locations (a depth map in which the number of the depth plane is shown by the level of gray). The recognized locations correspond to the structure of the source object.

The effect of noise was estimated by adding the background noise. The stability of the recognition of coordinates under noisy conditions was confirmed [11]. The grayscale images were analyzed as well; the edges of tetrahedron were painted in various colors and then transformed back to grayscale. According to [11], the convolution of the grayscale and color images became 30-40% asymmetric. Therefore, the recognition of locations in the grayscale images is less stable.

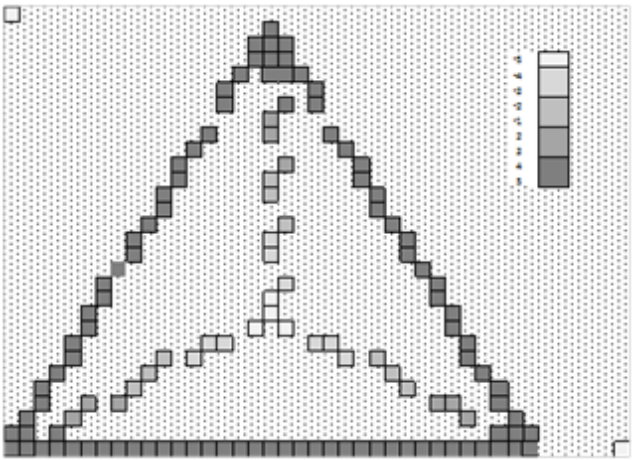

Fig. 6. Spatial structure of tetrahedron restored by convolution. Recognized depth is shown by levels of gray as in a depth map.

To improve the analysis method, the reference functions (3) can be considered as the scaling functions of a wavelet transform and thus the multiview wavelets can be built. It is important that the orthogonality is not a requirement for the scaling functions.

The difference between the rectangular unit pulse and the Haar function [15] is only in the locations of the edges of the pulse. Therefore, the reference functions (2) can be alternatively re-expressed as the linear combinations of the shifted Haar functions,

$$F_k(x) = \sum_{i=0}^{|k|-1} Haar\left(\frac{x - x_{kj}}{w_k} + \frac{1}{2}\right) \quad (4)$$

The 2D functions can be re-expressed in the similar manner. The corresponding multiview wavelets are the linear combinations of the shifted Haar wavelets [15].

The multiview scaling functions constructed this way are shown in Fig. 7. The corresponding multiview wavelets obtained using (1) are shown in Fig. 8.

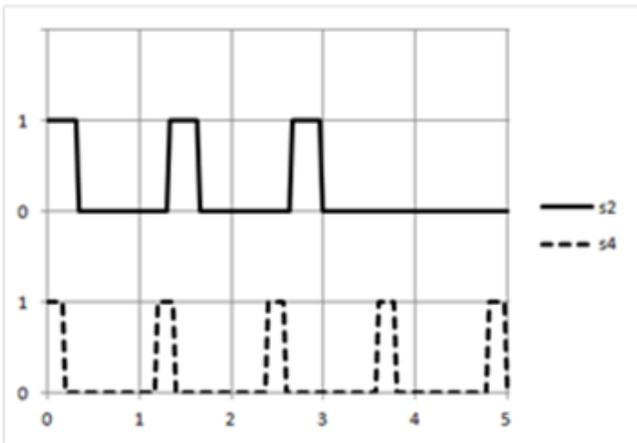

Fig. 7. Multiview scaling functions as functions of $x$-coordinate ($n = 2$ and $n = 4$).

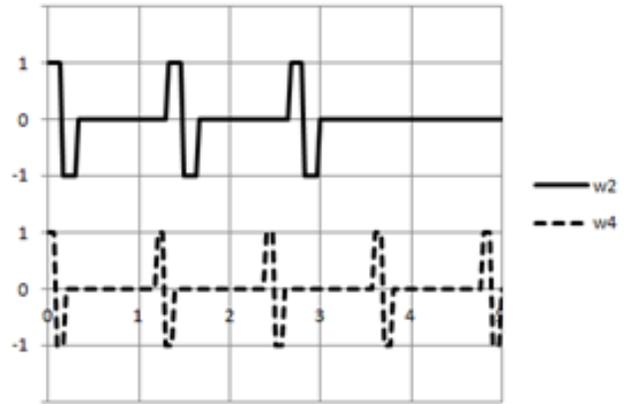

Fig. 8. Multiview wavelets as functions of $x$-coordinate ($n = 2$ and $n = 4$).

In the 2D case, the wavelets are the product of two 1D wavelets for $x$- and $y$-directions. An example is shown in Fig. 9 for $n = 2$.

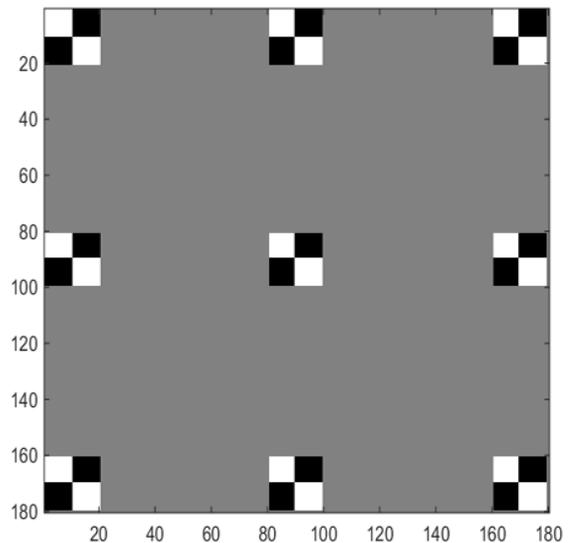

Fig. 9. 2D wavelet $n = 2$. (The pixels are numbered; one cell = 60 pixels.)

Using the wavelets Figs. 8, 9, the continuous wavelet transform can be performed for HPO and full-parallax multiview images (1D and 2D cases, resp.). An example of the wavelet analysis of the testing image Fig. 2(b) is shown in Fig. 10 for positive indexes $n$.

In Figs. 10(a) – (d), the displacement of the recognized point (which has the maximum wavelet transform and is the darkest in the graph therefore) along the left lower edge of tetrahedron (from the corner to the center, as expected) can be clearly seen. Similarly, all voxels of the testing tetrahedron are recognized, from a multiview image. For the discrete wavelet transform, the functions with numbers $2^n$ can be used, i.e., the 1st, 2nd, 4th, etc. functions. There is a good agreement in the restored locations. The wavelet analysis (CWT) using the proposed wavelets confirms that the recognized locations coincide with the coordinates of voxels of the original binary wireframe image.

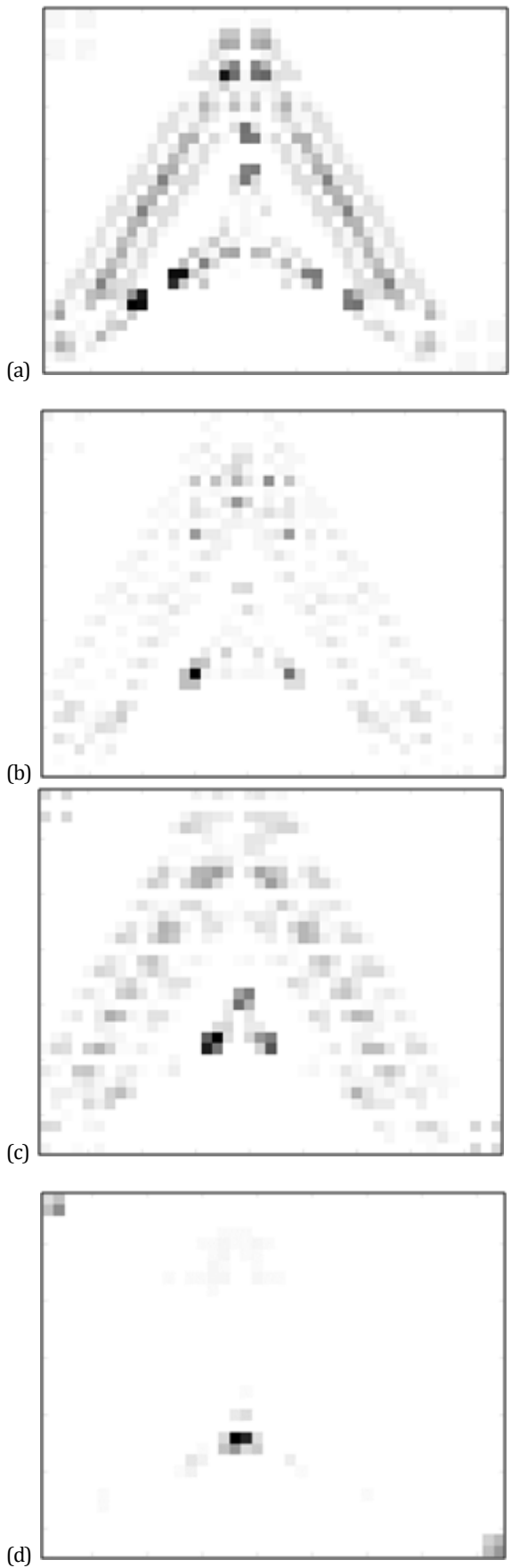

Fig. 10. 2D Wavelet analysis $n$ = 1, ..., 4 in (a), ..., (d), resp.

The proposed wavelet transform shows the characteristic features of the testing object. The spatial structure can be successfully extracted